\documentclass[12pt]{iopart}

\usepackage{iopams}  

\expandafter\let\csname equation*\endcsname\relax
\expandafter\let\csname endequation*\endcsname\relax

\usepackage{epsfig, amsmath, amssymb}
\newtheorem{theorem}{Theorem}

\newtheorem{assumption}{Assumption}
\usepackage{hyperref}
\usepackage{lineno}

\usepackage{xcolor}
\usepackage{cite}

\begin{document}

\title[Quantum LEGO Learning: A Modular Design Principle for Hybrid AI]{Quantum LEGO Learning: A Modular Design Principle for Hybrid Artificial Intelligence}

\author{Jun Qi$^{1}$, Chao-Han Huck Yang$^{2}$, Pin-Yu Chen$^{3}$, Min-Hsiu Hsieh$^{4}$, Hector Zenil$^{5}$, Jesper Tegner$^{6}$}

\address{1. School of Electrical and Computer Engineering, Georgia Institute of Technology, Atlanta, GA 30332, USA \\
2. NVIDIA Research, Santa Clara, CA 95051, USA		 \\
3. IBM Research, Yorktown Heights, NY 10598, USA		 \\
4. Hon Hai (Foxconn) Quantum Computing Research Center, Taipei, 114, Taiwan   \\ 
5. Biomedical Engineering and Imaging Sciences, King's College London, London
WC2R 2LS, UK                                             \\
6. Computer, Electrical, and Mathematical Sciences and Engineering Division, King
Abdullah University of Science and Technology, Thuwal 23955-6900, Saudi Arabia
}
\ead{jqi41@gatech.edu, jesper.tegner@kaust.edu.sa}
\vspace{10pt}


\begin{abstract}
Hybrid quantum-classical learning models increasingly integrate neural networks with variational quantum circuits (VQCs) to exploit complementary inductive biases. However, many existing approaches rely on tightly coupled architectures or task-specific encoders, limiting conceptual clarity, generality, and transferability across learning settings. In this work, we introduce Quantum LEGO Learning, a modular and architecture-agnostic learning framework that treats classical and quantum components as reusable, composable learning blocks with well-defined roles. Within this framework, a pre-trained classical neural network serves as a frozen feature block, while a VQC acts as a trainable adaptive module that operates on structured representations rather than raw inputs. This separation enables efficient learning under constrained quantum resources and provides a principled abstraction for analyzing hybrid models. We develop a block-wise generalization theory that decomposes learning error into approximation and estimation components, explicitly characterizing how the complexity and training status of each block influence overall performance. Our analysis generalizes prior tensor-network-specific results and identifies conditions under which quantum modules provide representational advantages over comparably sized classical heads. Empirically, we validate the framework through systematic block-swap experiments across frozen feature extractors and both quantum and classical adaptive heads. Experiments on quantum dot classification demonstrate stable optimization, reduced sensitivity to qubit count, and robustness to realistic noise.

\end{abstract}

\section{Introduction}
\label{sec1}

The rapid progress of quantum hardware has stimulated sustained interest in hybrid quantum–classical machine learning~\cite{biamonte2017quantum, schuld2015introduction, huggins2019towards}, in which variational quantum circuits (VQCs) are integrated with classical neural networks to form trainable models~\cite{cerezo2021variational, schuld2021effect, qi2025tensorhyper}. In the noisy intermediate-scale quantum (NISQ) regime~\cite{preskill2018quantum, bharti2022noisy}, VQCs are particularly attractive due to their flexible parameterization, compatibility with gradient-based optimization~\cite{cerezo2022challenges, holmes2022connecting}, and potential robustness to specific noise processes~\cite{resch2021benchmarking}. At the same time, their practical applicability remains fundamentally constrained by limited qubit counts, shallow circuit depths, and optimization pathologies such as barren plateaus and noise-induced gradient variance~\cite{mcclean2018barren, mitarai2018quantum, larocca2025barren}. As a result, standalone VQCs often struggle to model complex, high-dimensional data encountered in modern artificial intelligence, such as images, structured signals, and biological sequences~\cite{caro2022generalization, power_data}.

A widely adopted strategy to mitigate these limitations is to combine VQCs with strong classical feature extractors, such as convolutional neural networks~\cite{szegedy2017inception, he2016deep}, tensor-network models~\cite{chen2021end, yang2017tensor, qi2023exploiting}, or transformers~\cite{han2022survey}, thereby delegating most representational complexity to classical encoders while reserving quantum circuits for downstream adaptation. Although this paradigm has demonstrated promising empirical performance, existing analyses are typically tied to specific architectural choices~\cite{ostaszewski2021reinforcement, rapp2025reinforcement, ren2021comprehensive}, most notably tensor-network-based encoders or fixed feature maps. Consequently, it remains unclear why such hybrid systems generalize, which components drive performance gains, and how behavior scales across different classical backbones and quantum heads~\cite{qi2023theoretical}. This architectural coupling has also raised concerns regarding generality, interpretability, and the risk that observed improvements merely reflect classical inductive biases expressed through a quantum interface.

\begin{figure}[t]
\centerline{\epsfig{figure=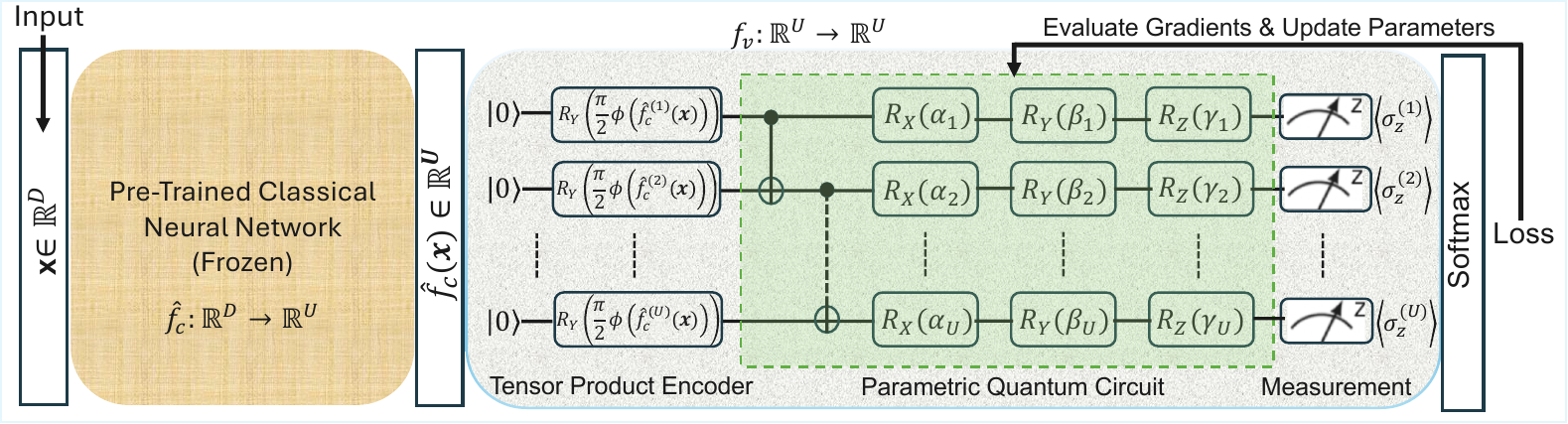, width=155mm}}
\caption{{\it Quantum LEGO Learning Architecture}. A pre-trained classical neural network (frozen) serves as the feature block, transforming the input $\textbf{x} \in \mathbb{R}^D$ into a structured embedding $\widehat{f}_{c}(\textbf{x}) \in \mathbb{R}^U$. This embedding is then encoded into a quantum state via a Tensor Product Encoder using single-qubit $R_{Y}(\frac{\pi}{2}\phi(\widehat{f}^{(u)}_{c}(\textbf{x})))$ rotations. A trainable VQC acts as the quantum block, applying layers of parameterized $R_{X}(\alpha_u)$, $R_{Y}(\beta_u)$, and $R_{Z}(\gamma_u)$ gates, highlighted by a green dashed box. Measurement of Pauli-Z observables $\langle \sigma_{z}^{(u)}\rangle$ produces classical outputs that are passed through a softmax layer for prediction. During training, only the quantum block parameters are updated, while the classical feature block remains fixed. This modular design exemplifies Quantum LEGO Learning, in which classical and quantum components serve as reusable, composable blocks.}
\label{fig:lego}
\end{figure}

In this work, as shown in Fig.~\ref{fig:lego}, we introduce Quantum LEGO Learning, a general compositional framework for hybrid quantum–classical models that explicitly separates learning into reusable, modular blocks. The central design principle is to treat a pre-trained classical neural network as a frozen feature block, while restricting all trainable parameters to a quantum adaptation block implemented by a VQC. This yields a hybrid operator of the form
\begin{equation}
\label{eq:lego}
f_{\rm lg} = \widehat{f}_{v} \circ f_{c}, 
\end{equation}
where $\widehat{f}_{c}$ is a fixed pre-trained encoder and $f_{v}$ contains the only trainable parameters. Unlike prior approaches~\cite{qi2023theoretical}, Quantum LEGO Learning is not tied to a specific encoder architecture or circuit family; instead, it provides a block-level abstraction that applies uniformly across convolutional networks, tensor networks, and other classical models.

This modularity enables a learning-theoretic analysis that cleanly decomposes generalization error into approximation, estimation, and optimization terms associated with individual blocks. Crucially, we show that once a sufficiently expressive classical encoder is frozen, the approximation error of the hybrid model becomes weakly dependent on the quantum dimension, while optimization stability improves due to the reduced parameter space seen by the VQC. These results generalize prior tensor-network-specific analyses and yield architecture-agnostic conditions under which quantum blocks can offer representational advantages over comparably sized classical heads. 

To empirically validate these insights, we conduct systematic block-swap experiments across multiple classical feature extractors (including ResNet variants~\cite{wen2020transfer} and tensor-train network (TTN)~\cite{oseledets2011tensor}) and across both quantum and classical trainable heads under matched parameter budgets. Experiments on quantum dot classification and genome TFBS prediction tasks, under noise-free simulation, realistic noise models, and real IBM quantum hardware~\cite{steffen2011quantum, shukla2020complete}, demonstrate consistent performance gains, stable optimization~\cite{martin2023barren}, and graceful degradation under noise. Together, these results position Quantum LEGO Learning as a principled and extensible framework for designing scalable, interpretable, and noise-resilient hybrid learning systems in the NISQ era~\cite{qi2025tensorhyper}.

\section{Results}

\subsection{Methodology: The Quantum LEGO Learning Framework}
Quantum LEGO Learning views hybrid quantum–classical models as composable assemblies of functional blocks, each contributing a specific transformation or inductive bias. A model is constructed by stacking two reusable components: (1) A pre-trained classical feature block, which is fully frozen during downstream training; and (2) A trainable quantum block, implemented as a VQC whose parameters are the only parameters updated during learning. This strict division of roles is a core characteristic of the LEGO framework: the classical block provides a stable, reusable representation, while the VQC acts as the sole adaptive module. We separately describe each component below.

\textit{Pre-Trained Classical Feature Block (Frozen)}. The classical block, denoted by $f_{c}$, is a neural network, such as ResNet-18/50 or a TTN, trained on a large source dataset $\mathcal{D}_{A}$. Its role is to generate high-level features that a shallow quantum circuit cannot easily learn from scratch. 

During downstream training, all parameters in this block are frozen. No gradient flows through this module, and no updates are made. The block maps raw inputs $\textbf{x} \in \mathbb{R}^{D}$ to lower-dimensional embeddings: 
\begin{equation}
\widehat{f}_{c}(\textbf{x}) \in \mathbb{R}^{U}, 
\end{equation}
where the dimension $U$ matches the number of qubit channels, this module serves as a frozen LEGO piece, providing stable, reusable representation power independent of the learning process, ensuring that all trainable capacity is concentrated in the quantum block.

\textit{Tensor Product Encoder (TPE)}. The TPE serves as the interface between the frozen classical block and the trainable quantum block. It converts the classical embedding into a quantum-ready state without introducing additional trainable parameters. Formally, given the input vector $\textbf{x}$, the TPE constructs a separable product state: 
\begin{equation}
\vert \bar{\textbf{x}} \rangle = \otimes_{u=1}^{U} R_{Y}\left(\frac{\pi}{2} \phi(\widehat{f}_{c}(\textbf{x})) \right) \vert 0 \rangle,
\end{equation}
where each qubit encodes its feature independently. Because the encoder contains no learnable parameters, the only adaptive component of the entire hybrid model is the VQC head.

\textit{Trainable Parametric Quantum Block (Only Trainable Component)}. The trainable parametric quantum block is the only part of the architecture that undergoes learning. Once the input has been transformed into a quantum state by the TPE, this block applies a sequence of adjustable quantum operations to extract task-relevant information. Each layer contains: (1) single-qubit rotations $R_{X}(\alpha_{u})$, $R_{Y}(\beta_{u})$, $R_{Z}(\gamma_{u})$, whose parameters $\alpha_{u}$, $\beta_{u}$, $\gamma_{u}$ are learnable; and (2) entangling gates (e.g., CNOT) that capture correlations across qubits. These operations in the quantum block are stacked at shallow depth to remain NISQ-compatible. 

During training, the quantum block is updated only via classical stochastic gradient descent with respect to the cross-entropy loss, while the classical block remains fixed. Notably, the gradients are estimated solely from repeated quantum measurements. This strict parameter isolation ensures that all adaptation is absorbed by the VQC, incredibly stabilizing training by preventing destructive interference between classical and quantum parameter spaces.

\textit{Measurement and Classification}. The VQC outputs expectation values $\langle \sigma_{z}^{(u)} \rangle$, $\forall u\in [U]$, which are passed to a classical softmax layer for prediction. Gradients are estimated via repeated sampling, and only the parameters within the VQC block receive updates, while the frozen classical block remains unchanged throughout training.

\subsection{Theoretical Results: Block-Wise Error Bounds}

We now present the theoretical foundations of Quantum LEGO Learning. Our analysis decomposes the expected risk of the hybrid model into approximation, estimation, and optimization errors~\cite{mohri2018foundations, qi2020analyzing}: 
\begin{equation}
\mathcal{L}(f_{\rm lg}) = \epsilon_{\rm app} + \epsilon_{\rm est} + \epsilon_{\rm opt}. 
\end{equation}

A key property of the LEGO architecture is that only the VQC block is trainable, while the classical block is fixed. This separation allows our error analysis to partition the contributions of both blocks clearly. In particular, unlike earlier analyses tied to TTN-VQC structures~\cite{qi2023theoretical}, our derivations apply to any frozen pre-trained encoder $\widehat{f}_{c}$ and any VQC structure $f_{v}$. More crucially, it is suitable for architectures in which the quantum block is the only trainable component. 

This enables block-wise complexity measures to replace architecture-specific arguments and yields general bounds on approximation, estimation, and optimization errors. The approximation error corresponds to the representation power, and the estimation and optimization errors are jointly related to the generalization capability. 

\textit{Approximation Error Bound}. The key theoretical insight is that the hybrid model's approximation error depends primarily on the complexity of the classical feature block and is independent of the number of qubits, as shown in Theorem~\ref{thm:app}.
\begin{theorem}[Approximation Error].
\label{thm:thm1}
\label{thm:app}
Let $\mathbb{F}_{c}$ be the functional class of the pre-trained model and $D_{A}$ the source dataset. Then, the approximation error
\begin{equation}
\label{eq:app}
\epsilon_{\rm app} = \mathcal{O}\left( \sqrt{\frac{\mathcal{C}(\mathbb{F}_c)}{\vert D_{A}\vert}} \right) + \mathcal{O}\left( \frac{1}{\sqrt{M}} \right), 
\end{equation}
where $\mathcal{C}(\cdot)$ is a complexity measure (e.g., empirical Rademacher complexity) and $M$ denotes the measurement count. 
\end{theorem}

Theorem~\ref{thm:thm1} suggests that increasing qubits does not improve approximation capacity, and more expressive classical pre-trained features significantly tighten the bound. This generalizes and strictly strengthens prior TTN-specific results, in which the upper bound relies heavily on the number of qubits. 

\textit{Estimation Error Bound}. The estimation error depends solely on the complexity of the quantum block and the amount of task-specific labeled data. 

\begin{theorem}[Estimation Error]. 
\label{thm:thm2}
Let $\mathbb{F}_{v}$ be the functional class of the VQC, and $D_{B}$ be the target dataset. Then, the estimation error 
\begin{equation}
\epsilon_{\rm est} = \mathcal{O}\left(	\sqrt{\frac{\mathcal{C}(\mathbb{F}_v)}{\vert D_{B} \vert}}	\right), 
\end{equation}
where $\mathcal{C}(\cdot)$ is a complexity measure (e.g., empirical Rademacher complexity). 
\end{theorem}

Theorem~\ref{thm:thm2} shows that if the downstream dataset is small, using a pre-trained encoder is crucial. The VQC complexity scales with the number of qubits and the depth of its layers, consistent with empirical observations.

\textit{Optimization Error Bound}. The frozen classical block dramatically simplifies optimization by removing parameter interactions and reducing curvature variation. Under the assumptions of approximate linearity and gradient boundedness in Assumptions~\ref{ass1} and~\ref{ass2}, we can derive the upper bound on the optimization error $\epsilon_{\rm opt}$ in Theorem~\ref{thm:thm3}. 

\begin{assumption}[Approximate Linearity]. 
\label{ass1}
Let $D_{B} = \{\textbf{x}_1, \textbf{x}_2, ..., \textbf{x}_{\vert D_B \vert}\}$ be the target dataset. There exists a $f_{lg} = f_{v} \circ f_{c}$ with a VQC's parameter vector $\boldsymbol{\theta}$, denoted as $f_{v}(\textbf{x}_n; \boldsymbol{\theta})$. Then, for two constants $\beta$, $L$ and a first-order gradient $\boldsymbol{\delta}$, we have 
\begin{equation}
\sup\limits_{\boldsymbol{\delta}} \frac{1}{\vert D_B\vert} \sum\limits_{n=1}^{\vert D_{B} \vert} \lVert \nabla_{\boldsymbol{\theta}}^{2} f_{\rm lg}(\textbf{x}_{n}; \boldsymbol{\theta} + \boldsymbol{\delta})	\rVert_{2}^{2} \le \beta^{2}, 
\end{equation} 
and 
\begin{equation}
\sup\limits_{\boldsymbol{\delta}} \frac{1}{\vert D_B\vert} \sum\limits_{n=1}^{\vert D_B \vert} \lVert \nabla_{\boldsymbol{\theta}} f_{\rm lg}(\textbf{x}_n; \boldsymbol{\theta}) \rVert_{2}^{2} \le L^2. 
\end{equation}
\end{assumption}

\begin{assumption}[Gradient Bound]. 
\label{ass2}
For the first-order gradient $\boldsymbol{\theta}$ of the VQC's parameters in the training process, we assume that $\lVert \boldsymbol{\delta} \rVert_{2}^{2} \le R$ for a constant R. 
\end{assumption}

\begin{theorem}[Optimization Error]. 
\label{thm:thm3}
With a learning rate 
\begin{equation}
\eta = \frac{1}{T} \frac{R}{\sqrt{L^2 + \beta^2 R^2}}, 
\end{equation}
we obtain 
\begin{equation}
\epsilon_{\rm opt} \le \beta R^{2} + R\sqrt{\frac{L^{2} + \beta^2 R^2}{T}}. 
\end{equation}
\end{theorem} 

Theorem~\ref{thm:thm3} suggests that the upper bound on $\epsilon_{\rm opt}$ is controlled by the constant factor $R$ for the supremum value $\lVert \boldsymbol{\delta} \rVert_{2}^{2}$. Compared with the prior TTN-VQC model with the Polyak–Łojasiewicz condition to guarantee a small optimization error, the Quantum LEGO Learning ensures an upper bound related to the iteration step $T$ and the constant $R$ for the supremum value of $\lVert \boldsymbol{\delta} \rVert_{2}^{2}$. To reduce the upper bound of $\epsilon_{\rm opt}$, a larger $T$ and a small $R$ are expected to achieve a small learning rate $\eta$. 

\subsection{Noise-Resilience of Quantum LEGO Learning}

Quantum noise, arising from decoherence, gate fidelity, and measurement uncertainty, poses a central challenge for VQCs in the NISQ era. A distinguishing advantage of the Quantum LEGO Learning framework is that it naturally improves noise-resilience by freezing the classical feature block and isolating all trainable capacity in a VQC head. This isolation simplifies the optimization landscape, reduces the propagation of noise-induced gradients, and enables theoretical guarantees that would not hold in monolithic hybrid models. 

\textit{Structural Noise Mitigation via Block Freezing}. In traditional hybrid models, both classical and quantum components are trainable, so quantum noise affects the entire gradient flow. Noisy measurements perturb gradients that backpropagate through the classical block, amplifying noise and destabilizing optimization. In contrast, Quantum LEGO Learning disconnects the classical block from gradient flow. It suggests that only the VQC block receives noisy gradients, while the feature block produces stable embeddings that remain unaffected by noise during training. 

\textit{Noise-Buffered Optimization Dynamics}. Given the clean measurement operator $\boldsymbol{\sigma}_{z} = [\langle \sigma_{z}^{(1)} \rangle, \langle \sigma_{z}^{(2)} \rangle, ..., \langle \sigma_{z}^{(U)} \rangle]^{\top}$, we define a noisy measurement operator $\boldsymbol{\tilde{\sigma}}_{z}$ as:
\begin{equation}
\boldsymbol{\tilde{\sigma}}_{z} = \boldsymbol{\sigma}_{z} + \boldsymbol{\xi}, \hspace{2mm} \mathbb{E}[\boldsymbol{\xi}] = 0, \hspace{2mm} \mathbb{E}[\lVert \boldsymbol{\xi} \rVert_{2}^{2}] = \tau^{2},
\end{equation}
where the perturbation vector $\boldsymbol{\xi}$ relies on the VQC parameters $\boldsymbol{\theta}$. Since gradients flow only through the VQC, for a loss function $\mathcal{L}(\boldsymbol{\theta})$, at epoch $t+1$, the optimization update of VQC parameters $\boldsymbol{\theta}$ becomes:
\begin{equation}
\boldsymbol{\theta}_{t+1} = \boldsymbol{\theta}_{t} - \eta (\nabla_{\boldsymbol{\theta}}\mathcal{L}\left(\boldsymbol{\theta}_t) + \nabla_{\boldsymbol{\theta}}\boldsymbol{\xi}_{t}\right). 
\end{equation}

Because the classical block is non-trainable, the error term does not propagate into earlier layers. As a result, the cumulative noise after $T$ iterations satisfies: 
\begin{equation}
\mathbb{E}\left[ \left\lVert \sum\limits_{t=1}^{T} \eta \nabla_{\boldsymbol{\theta}} \boldsymbol{\xi}_t \right\rVert_{2}^{2} \right] = \eta^2 T \sigma^2, 
\end{equation}
which is instead of the $\mathcal{O}(\eta^2 T L^{2}_{v} \sigma^2)$ amplification found in hybrid models, where the classical block contributes to the VQC Lipschitz constant $L_{v}$. Given the classical model's Lipschitz constant $L_{c}$, because $L_{v} < L_{c}$, LEGO learning provides tight variance control. 

\textit{Theoretical Guarantee of Noise-Tolerant Convergence}. We extend the optimization bound in Theorem~\ref{thm:thm3} to include measurement noise. 

\begin{theorem}
\label{thm:thm4}
Under the assumptions of approximate linearity (Assumption~\ref{ass1}), gradient boundedness (Assumption~\ref{ass2}), and additive measurement noise with variance $\tau^2$, the optimization error of the Quantum LEGO model after $T$ iterations satisfies: 
\begin{equation}
\epsilon_{\rm opt, noise} \le \beta R^2 + \frac{R \sqrt{L^2 + \beta^2 R^2}}{T} + \eta R\tau \sqrt{T}. 
\end{equation}
\end{theorem}

In theorem~\ref{thm:thm4}, the first two terms match the noiseless optimization error of the LEGO learning, and the third term $\eta R\tau \sqrt{T}$ refers to a noise-induced penalty. Because the VQC block is only trainable, the term $R = \lVert \boldsymbol{\delta} \rVert_{2}$ is small (no classical parameter updates), making the noise term smaller than in models with trainable classical encoders. 

Thus, Quantum LEGO Learning is guaranteed to accumulate strictly less noise-induced optimization error than any architecture in which both classical and quantum blocks are updated. Moreover, the approximation error term remains the same as Eq. (\ref{thm:app}) in Theorem~\ref{thm:thm1}, which provides a natural robustness mechanism absent from architectures that rely on expressive quantum feature maps. 

\textit{Stability of the VQC Under Noise}. Quantum LEGO Learning stabilizes the VQC against noise in three ways: (i) Only the trainable VQC block needs expressive capacity, and the frozen classical block handles feature complexity. This keeps the quantum circuit depth low, decreasing exposure to noise channels such as amplitude damping or depolarization; (ii) Noise does not accumulate through backpropagation across many layers, as happens when classical blocks are trainable; (iii) The frozen classical block ensures the input distribution into the VQC remains stable across training. This prevents noise-induced shifts in the optimization path.

\subsection{Implementation of Gradient Descent on Quantum Hardware}

Training hybrid quantum-classical models on real quantum hardware requires careful handling of noisy measurements, hardware-specific gate constraints, and limited coherence times. Quantum LEGO Learning is expressly designed to simplify this process by ensuring that only the VQC block is trainable, while the classical block is entirely frozen. This architectural choice enables gradient-based optimization on hardware in a streamlined manner, since all learnable parameters reside in a compact, shallow quantum circuit. 

\textit{Parameter-Shift Gradient Estimation (Trainable Only in VQC)}. Because the classical feature block is fixed, the gradient descent requires only with respect to the VQC parameters $\boldsymbol{\theta} = \{ \alpha_u, \beta_u, \gamma_u \}_{u=1}^{U}$. Then, gradients are computed using the parameter-shift rule. For any trainable quantum gate of the form: 
\begin{equation}
U(\boldsymbol{\theta}) = \exp(-i\boldsymbol{\theta}\sigma_{z}), 
\end{equation}
where $\sigma_{z}$ is a Pauli-Z operator, the exact gradient of an expectation value is given by: 
\begin{equation}
\frac{\partial}{\partial \boldsymbol{\theta}} \langle f_{\rm lg}(\boldsymbol{\theta}) \rangle = \frac{1}{2} \left[ \left\langle f_{\rm lg}(\boldsymbol{\theta} + \frac{\pi}{2}) \right\rangle - \left\langle f_{\rm lg}(\boldsymbol{\theta} - \frac{\pi}{2}) \right\rangle	\right]. 
\end{equation}

In Quantum LEGO Learning, this rule applies to every trainable quantum parameter, since the classical encoder has no differentiable parameters. This simplifies hardware implementation: for each parameter, only two forward quantum evaluations are needed. 

\textit{Shot-Based Stochastic Gradients}. Each expectation value required by the parameter-shift rule is approximated using repeated quantum measurements (``shots"). Given $M$ shots per shift evaluation, the estimator is: 
\begin{equation}
\frac{\partial}{\partial \boldsymbol{\theta}} \langle f_{\rm lg}(\boldsymbol{\theta}) \rangle \approx \frac{1}{2} \left[ \bar{f}_{\rm lg}(\boldsymbol{\theta} + \frac{\pi}{2}) - \bar{f}_{\rm lg}(\boldsymbol{\theta} - \frac{\pi}{2}) \right], 
\end{equation}
where each $\bar{f}_{\rm lg}$ has sampling noise of order $\mathcal{O}(\frac{1}{\sqrt{M}})$. Because no classical parameters are updated, all stochasticity in training arises solely from quantum measurement noise, consistent with the analysis in the approximation and estimation error bounds. 

\textit{Hardware-Aware Learning Rate and Noise Accumulation}. On NISQ devices, gradient noise increases with circuit depth and shot-count variability. The LEGO architecture mitigates this by: (i) Keeping the VQC shallow, limiting noise accumulation; (ii) Reducing parameter coupling, because classical weights do not shift during training; and (iii) Enabling stable step-size schedules, where the learning rate is chosen as in Theorem~\ref{thm:thm3}.

\subsection{Empirical Results of Quantum Dot Classification}

\begin{figure}
\centerline{\epsfig{figure=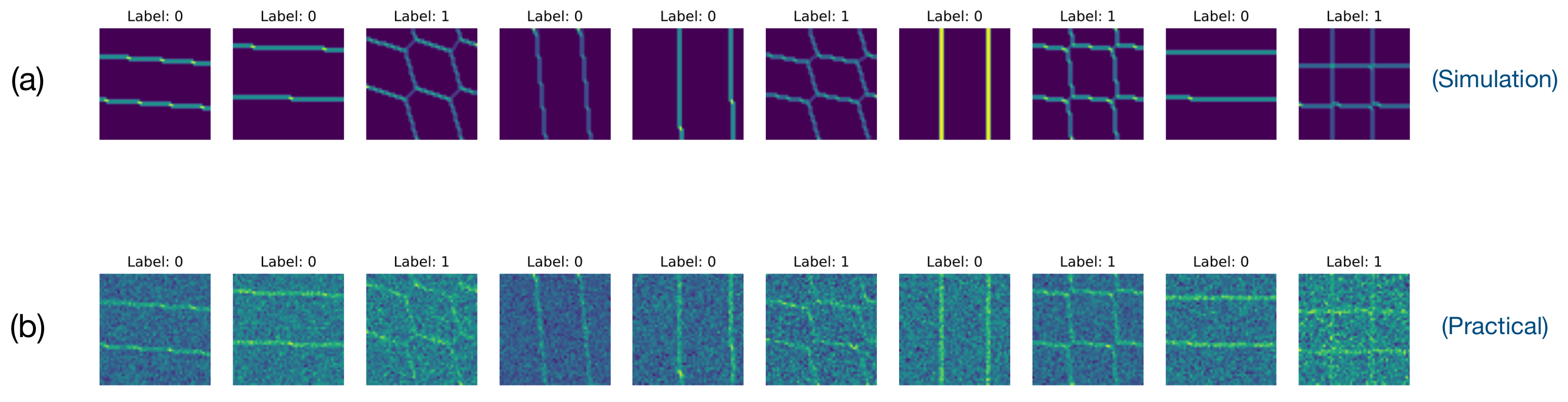, width=140mm}}
\caption{\textbf{Illustration of single and double quantum dot charge stability diagrams}. (a) labeled clean charge stability diagrams containing the transition lines without noise; (b) labeled noisy charge stability diagrams mixed with realistic noise effects on the transition lines. Label:$0$ and Label:$1$ denote charge stability diagrams of single and double quantum dots. By detecting transition lines using the QML approach, we aim to determine whether the charge stability diagram corresponds to single- or double-quantum-dot behavior. In particular, we use noiseless data to evaluate the models' representational power and noisy data to assess their generalization power.}
\label{fig:dot}
\end{figure}

\textit{Experimental Protocol}. To empirically validate the proposed Quantum LEGO Learning framework, we evaluate its generalization performance on the quantum dot classification task~\cite{czischek2021miniaturizing, kalantre2019machine, ziegler2023tuning}. As shown in Fig.~\ref{fig:dot}, this dataset comprises clean and noisy $50\times 50$ stability diagrams, labeled as single- or double-dot configurations~\cite{gualtieri2025qdsim}. Throughout all experiments, the classical feature block remains entirely frozen, and only the parameters of the VQC block are trained, providing clean isolation of quantum adaptation and directly testing theoretical predictions regarding approximation-error independence, optimization stability, and noise resilience. 

We conduct four groups of experiments: (1) Generalization capability of LEGO architecture across pre-trained ResNet18/ResNet50 blocks; (2) Block-swap experiments: replacing ResNet18/ResNet50 with pre-trained TTN feature blocks; (3) Quantum-noisy simulations with realistic depolarization, dephasing, and readout errors; and (4) Real-hardware validation on the IBM Heron quantum processor. All experiments follow a unified training and evaluation protocol. The VQC parameters are optimized using the Adam optimizer with a learning rate of $0.001$, and cross-entropy loss~\cite{mao2023cross} is employed for all classification experiments. Since the classical feature block remains frozen, gradient updates are computed solely with respect to the VQC parameters, ensuring stable, well-conditioned optimization.

\begin{figure}
\centerline{\epsfig{figure=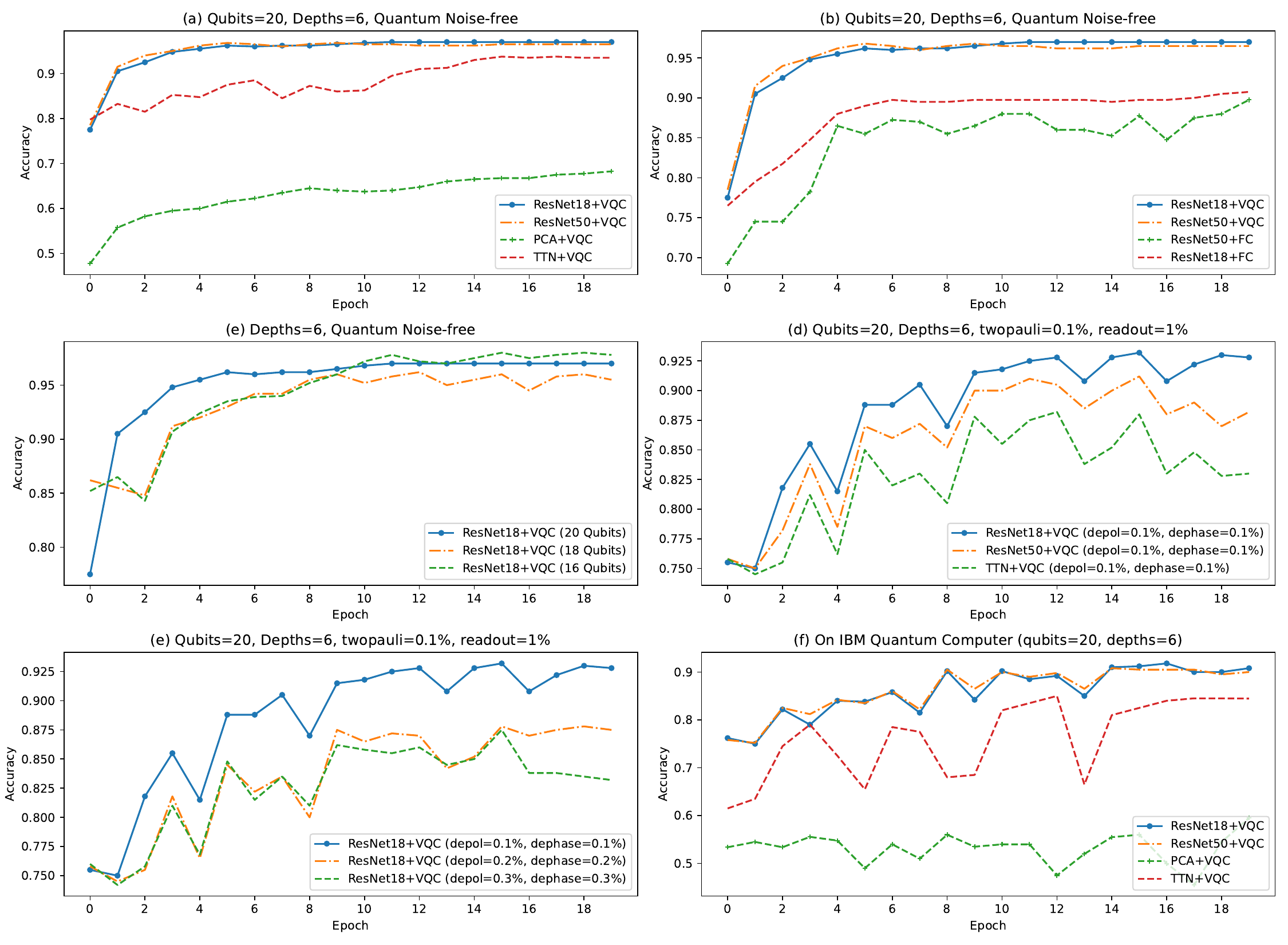, width=155mm}}
\caption{\textbf{Empirical evaluation of Quantum LEGO Learning on quantum dot classification}. Classification accuracy over training epochs under different hybrid architectures, noise conditions, and deployment settings. (a) Noise-free simulation with 20 qubits and 6 VQC layers, comparing ResNet18/50 + VQC with PCA + VQC and TTN + VQC; (b) comparison between quantum (VQC) and classical (FC) heads trained on identical frozen ResNet features; (c) sensitivity to qubit count under noise-free simulation, showing stable performance as qubit number decreases; (d) performance under realistic quantum noise (depolarizing, dephasing, two-qubit Pauli, and readout errors); (e) robustness of ResNet18 + VQC under increasing single-qubit noise strength; (f) execution on the IBM Heron quantum processor (20 qubits, depth 6), demonstrating practical robustness.}
\label{fig:res}
\end{figure}

\textit{Generalization under Noise-Free Quantum Simulation}. We first evaluate Quantum LEGO Learning under noise-free simulation with 20 qubits and 6 VQC layers, as summarized in Fig.~\ref{fig:res} (a). LEGO models built on ResNet18+VQC and ResNet50+VQC exhibit rapid convergence and achieve the highest classification accuracy among all architectures, stabilizing around $97\%$ accuracy within a few training epochs. The performance gap between ResNet18 and ResNet50 is marginal, indicating that once a sufficiently expressive frozen feature block is provided, the VQC primarily functions as an adaptive decision module rather than learning hierarchical representations from scratch.

In contrast, TTN+VQC achieves slightly lower but still competitive accuracy $(93–94\%)$. While the TTN offers a structured low-rank representation that substantially outperforms simple baselines (e.g., PCA+VQC), its expressivity remains more limited than that of deep convolutional encoders. The observed accuracy gap between TTN-based and ResNet-based LEGO models is consistent with the block-wise approximation-error analysis, which explicitly accounts for the complexity of the frozen classical block.

Importantly, all LEGO configurations exhibit smooth, monotonic training curves with no signs of optimization instability or barren plateau behavior, even at 20 qubits and depth 6. This confirms that constraining all trainable parameters to the VQC block yields a well-conditioned optimization problem in practice.

\textit{Quantum versus Classical Heads with Identical Frozen Features}. To isolate the contribution of the trainable quantum block, Fig.~\ref{fig:res} (b) compares ResNet18/50+VQC with ResNet18/50+FC classical heads trained on identical frozen features. Across both backbones, LEGO models with quantum heads consistently outperform their FC counterparts in both convergence speed and final accuracy. While classical FC heads improve during training, their performance saturates at a noticeably lower level, particularly for ResNet18.

This clean comparison validates a central LEGO-learning claim: the performance gain arises from the adaptive quantum block, rather than from additional classical representational power. Since both models operate on the same frozen feature space, the superior performance of ResNet+VQC highlights the ability of shallow quantum circuits to capture correlation structures that are inefficient for classical linear heads under comparable parameter budgets.

\textit{Sensitivity to Qubit Count}. Fig.~\ref{fig:res} (c) further examines the sensitivity of LEGO models to the number of qubits under noise-free simulation. ResNet18+VQC maintains stable performance as the qubit count is reduced from 20 to 18 and 16, with only minor variations in final accuracy. This observation empirically supports the theoretical prediction that the frozen classical block dominates the approximation error and exhibits weak dependence on the quantum dimension, enabling strong performance with relatively small quantum circuits.

\textit{Robustness under Realistic Quantum Noise}. We next introduce realistic quantum noise, including depolarizing, dephasing, two-qubit Pauli, and readout errors, as shown in Fig.~\ref{fig:res} (d)–(e). Under moderate noise levels, ResNet18/50+VQC models retain stable convergence and high accuracy, with only gradual performance degradation as noise strength increases. TTN+VQC exhibits similar qualitative behavior but degrades more rapidly, consistent with its lower baseline expressivity.

These results confirm that Quantum LEGO models degrade gracefully rather than catastrophically under noise, empirically validating the noise-resilience properties implied by the block-wise error decomposition and the isolation of quantum adaptation to a shallow, trainable circuit.

\textit{Real-Hardware Validation on IBM Heron}. Finally, Fig.~\ref{fig:res} (f) reports execution on the IBM Heron quantum processor with 20 qubits and depth 6. Despite hardware noise and limited shot budgets, ResNet18/50+VQC achieves stable accuracies around $90–92\%$, substantially outperforming both standalone VQC and TTN+VQC models. In contrast, pure VQC models remain near-chance level, underscoring the critical importance of frozen, pre-trained classical blocks for practical quantum learning on current hardware.

Taken together, these results demonstrate that Quantum LEGO Learning is architecture-agnostic, noise-resilient, and practically viable on real quantum hardware. The consistent superiority of VQC heads over classical FC heads, combined with weak dependence on qubit count and strong robustness to noise, provides compelling empirical support for the proposed theoretical framework.

\subsection{Empirical Results of Genome TFBS Prediction}
To further evaluate the proposed Quantum LEGO Learning framework on structured, high-dimensional biological data, we consider a genome transcription factor binding site (TFBS)~\cite{tompa2005assessing, ou2018motifstack} prediction task and deploy the hybrid models on the IBM Heron quantum processor. The task is to predict whether a given DNA sequence contains a binding site for the transcription factor JunD, a member of the activator protein-1 family that plays a critical role in gene regulation. 

\begin{figure}
\centerline{\epsfig{figure=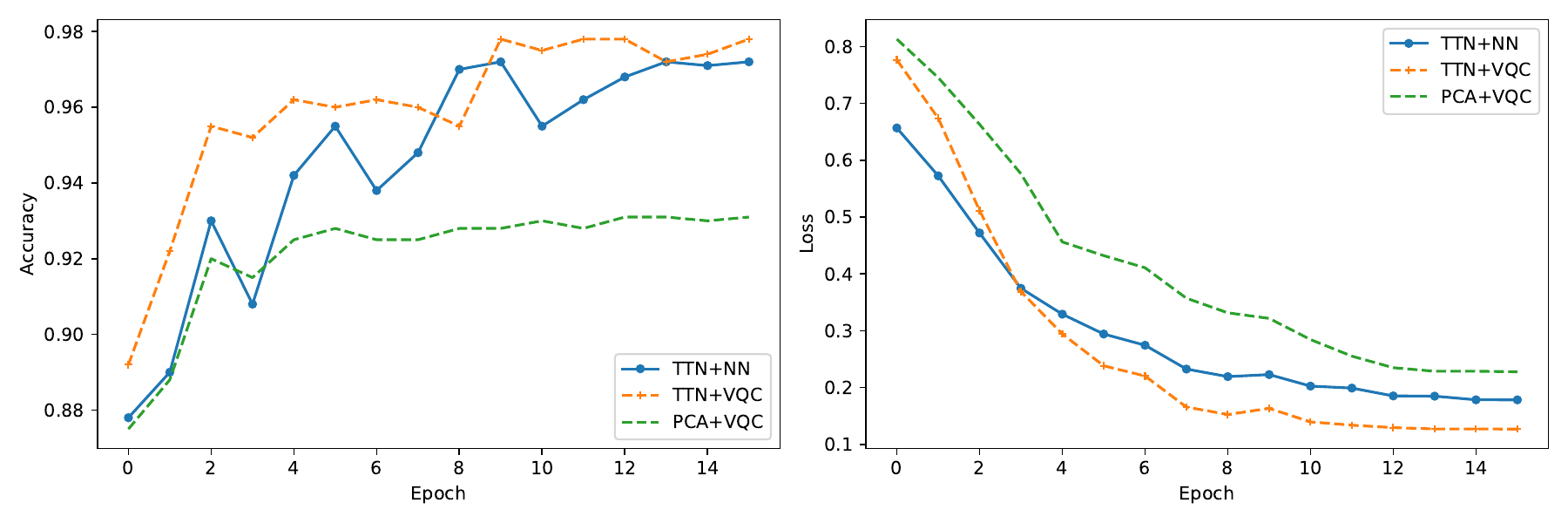, width=155mm}}
\caption{\textbf{Empirical results of genome TFBS prediction on IBM quantum hardware}. Classification accuracy (left) and cross-entropy loss (right) as a function of training epochs for genome TFBS prediction using hybrid Quantum LEGO models executed on the 156-qubit IBM Heron r2 processor. We compare a hybrid TTN+VQC model against a classical TTN+NN baseline and a PCA+VQC model, in which the pre-trained TTN feature block is frozen, and only the VQC parameters are optimized during training. Results show that Pre-TTN+VQC achieves higher accuracy and lower loss, demonstrating improved generalization and robustness under realistic quantum hardware conditions.}
\label{fig:tf_res}
\end{figure}

The dataset is derived from genome-wide binding experiments that identify JunD binding locations across all $22$ human chromosomes. Each chromosome is segmented into sequences of length $101$ bases, and each segment is labeled to indicate the presence or absence of a JunD binding site. Accordingly, using one-hot encoding over the $4$ nucleotide bases $\{A, C, G, T\}$, each input sequence is represented as a $404$-dimensional binary feature vector. 

Following the Quantum LEGO paradigm, we employ a TTN as a pre-trained classical feature block, resulting in the TTN+VQC hybrid model. The TTN is pre-trained on related image and speech processing tasks and remains frozen during downstream training. The quantum block is implemented as a shallow-depth variational quantum circuit with $8$ qubits and is the only trainable component of the hybrid architecture. We compare this model against a standalone VQC baseline and a classical counterpart, TTN+NN, where a neural network head replaces the VQC with a matched parameter budget. 

Figure~\ref{fig:tf_res} reports the classification accuracy and cross-entropy loss over training epochs obtained from execution on the IBM Heron processor. The results show that TTN+VQC consistently achieves higher accuracy and lower loss than both the standalone VQC and the classical TTN+NN model. Notably, the performance gap is more pronounced under hardware noise, indicating that the frozen TTN feature block helps stabilize optimization and mitigate the impact of quantum noise on generalization. These findings further support the central claim of Quantum LEGO Learning: composing a robust, pre-trained classical block with a shallow, trainable VQC yields improved generalization and hardware robustness for realistic quantum machine learning tasks. 

\section{Discussion}
This work introduces Quantum LEGO Learning as a modular and learning-theoretic framework for hybrid quantum–classical models, motivated by the observation that practical quantum machine learning systems increasingly rely on strong classical components to compensate for limited quantum resources. By explicitly decomposing hybrid models into a frozen classical feature block and a trainable quantum adaptation block, the framework provides both conceptual clarity and practical benefits in terms of optimization stability, generalization behavior, and robustness to noise.

From an AI perspective, a central insight of Quantum LEGO Learning is that joint end-to-end training is neither necessary nor desirable in many hybrid settings. Freezing a pre-trained classical block substantially reduces the set of adequate hypotheses explored during training, leading to smoother optimization landscapes and more stable gradients for the VQC. Our block-wise error analysis formalizes this intuition by isolating the approximation burden of the classical encoder from the estimation and optimization behavior of the quantum block. This decomposition explains why effective hybrid learning can be achieved even with shallow circuits and modest qubit counts, and why performance often exhibits weak sensitivity to quantum dimension once classical representations are fixed.

The empirical results on quantum dot classification reinforce these theoretical findings. Across multiple classical encoders, quantum heads consistently outperform classical heads trained on identical frozen features, while exhibiting smooth convergence and robustness under realistic noise models and on real quantum hardware. Importantly, these gains do not arise from end-to-end quantum processing, but from using VQCs as adaptive correlation modules layered atop strong classical representations. This observation aligns with a broader AI principle: near-term advantages of quantum learning are most likely to be realized in modular hybrid regimes, where quantum components complement rather than replace classical models.

At the same time, we emphasize that Quantum LEGO Learning does not claim universal superiority for quantum computation. The framework explicitly clarifies when and why quantum blocks are beneficial and when they may not be. If a classical encoder already linearizes the task, or if classical heads of comparable size efficiently capture relevant correlations, then quantum advantages may diminish. Similarly, while freezing classical blocks improves stability, partial fine-tuning may offer additional flexibility at the cost of increased optimization complexity. By making these trade-offs explicit, the block-based perspective helps prevent overclaiming and supports more transparent evaluation of quantum contributions.

More broadly, Quantum LEGO Learning provides a unifying abstraction that encompasses and extends several existing hybrid approaches, including tensor-network-guided quantum models and encoder–quantum-head architectures. By framing these methods within a common block-based viewpoint, the framework exposes fundamental design trade-offs and enables principled extension to new architectures, datasets, and noise regimes. Future directions include extending the analysis to adaptive or Bayesian classical blocks, integrating continual or meta-learning strategies, and exploring structured quantum heads tailored to specific correlation classes.

In summary, Quantum LEGO Learning reframes hybrid quantum–classical modeling as a compositional learning problem in which classical and quantum components play complementary, well-defined roles. By combining general theory with controlled empirical validation, including hardware experiments, this work contributes a scalable, interpretable, and practically grounded foundation for hybrid learning systems at the intersection of quantum computing and artificial intelligence. 

\section{Methods}

\subsection{TTN and TTN+VQC Architectures}

Tensor-Train Networks (TTNs), also known as Matrix Product States (MPS), are a class of tensor network models widely used in machine learning and quantum physics for representing high-dimensional functions in a parameter-efficient manner. A TTN factorizes a high-order tensor into a sequence of lower-order tensors, typically matrices or three-way tensors, connected in a linear, chain-like structure~\cite{oseledets2011tensor}. This decomposition enables TTNs to capture structured correlations in high-dimensional data while controlling model complexity through the tensor bond dimensions. 

When combined with a VQC, the resulting hybrid model is called TTN+VQC. In this architecture, the TTN serves as a classical preprocessing block that transforms the input data into a low-dimensional representation suitable for quantum encoding. The TTN output is subsequently encoded into a quantum state using a tensor-product encoder, in which each component of the embedding controls a single-qubit rotation. In particular, within the proposed Quantum LEGO Learning framework, TTN+VQC constitutes a specific instantiation of a compositional hybrid model. During downstream training, only the VQC parameters are optimized, while the TTN remains fixed. This design isolates quantum adaptation to the VQC, simplifies optimization, and allows block-wise analysis of approximation and generalization behavior. 

\subsection{Pre-trained ResNet Models}

Residual Networks (ResNets) are widely used convolutional neural network architectures that mitigate vanishing gradients through residual skip connections. In this work, we use pre-trained ResNet18 and ResNet50 as frozen feature extractors within the Quantum LEGO Learning framework. 

In particular, ResNet18 and ResNet50 differ primarily in depth and parameter count. ResNet18 contains approximately 11 million parameters, offering a lightweight architecture suitable for computationally constrained settings or tasks requiring fast inference, whereas ResNet50, with approximately 25 million parameters, provides a deeper, more expressive model capable of capturing richer hierarchical representations, albeit at increased computational cost. 

Through all experiments, these pre-trained models are used strictly as frozen encoders: no parameters of the ResNet feature block are updated during training. This isolation ensures that all trainable components reside in the quantum module (VQC), enabling a clean assessment of how the quantum block adapts downstream of a fixed classical representation. The use of pre-trained ResNets aligns with standard transfer learning practices and allows direct comparison between classical and quantum heads under equivalent feature inputs.

\subsection{Principal Component Analysis}

Principal Component Analysis (PCA) is a classical technique for dimensionality reduction that identifies directions of maximal variance in high-dimensional data. By projecting data onto a small number of orthogonal principal components, PCA yields compact representations that retain the most informative variation in the original dataset. In this work, PCA serves as an alternative classical feature extractor within the LEGO Learning framework. The method is beneficial when the dimensionality of the raw data is large relative to the number of available quantum channels in the VQC.

We use PCA to reduce high-dimensional classical inputs into feature vectors with dimensionality equal to the number of qubits in the quantum block. This enables direct comparison between PCA+VQC and ResNet/TTN-based LEGO models, isolating the effect of representational strength in the pre-trained feature block. PCA thereby provides a lightweight baseline feature transformation and highlights the interplay between classical dimensionality reduction and quantum expressivity.

\subsection{Cross-Entropy Loss Function}

For all classification tasks considered in this work, we employ the cross-entropy loss as the training objective. Given a dataset $\{(\textbf{x}_n, y_n)\}^{N}_{n=1}$ with input samples $\textbf{x}_{n}$ and corresponding class labels $y_n \in \{1, 2, ..., C\}$, the hybrid model outputs a probability distribution $\widehat{\textbf{p}}_{n} \in \mathbb{R}^{C}$ over $C$ classes after measurement of the quantum circuit and application of a softmax layer. The cross-entropy loss is defined as: 
\begin{equation}
\mathcal{L}_{\rm ce} = -\frac{1}{N} \sum\limits_{n=1}^{N}\log \widehat{p}_{n, y_{n}}, 
\end{equation}
where $\widehat{p}_{n, y_n}$ denotes the predicted probability assigned to the true class of sample $n$. 

In our proposed Quantum LEGO Learning framework, the cross-entropy loss is used to evaluate the performance of both standalone VQC models and Quantum LEGO Learning models. During optimization, gradients of $\mathcal{L}_{\rm ce}$ are computed exclusively with respect to the parameters of the VQC. In contrast, the parameters of the pre-trained classical feature block remain frozen. This design isolates the optimization process to the quantum block and ensures stable training by preventing interference between classical representation learning and quantum adaptation.

\subsection{IBM Quantum Hardware: Heron r2 Processor}
All real-device experiments in this work were executed on IBM’s Heron r2 quantum processor~\cite{shukla2020complete}, a state-of-the-art superconducting quantum processing unit~\cite{steffen2011quantum} equipped with 156 physical transmon qubits. The Heron r$2$ is part of IBM’s next-generation Heron family, designed to deliver substantial improvements in qubit connectivity, cross-talk suppression, and two-qubit gate fidelities relative to the preceding Eagle-class processors. This system employs a tunable-coupler architecture that dynamically optimizes qubit–qubit interactions to enable high-fidelity entangling operations. Typical two-qubit gate error rates are below $10^{-3}$, and the processor achieves a CLOPS throughput exceeding $2.0\times 10^{5}$, reflecting both high computational efficiency and stability. With its combination of mid-scale qubit capacity and low intrinsic noise, the Heron r$2$ platform is particularly suitable for benchmarking variational quantum algorithms in the NISQ regime.

\subsection{Proof of Theorem~\ref{thm:thm1}}

Let $\mathbb{F}_c$ be the hypothesis class of the classical encoder $f_c$, with complexity measure $\mathcal{C}(\mathbb{F}_c)$ (e.g., empirical Rademacher complexity). For a pre-trained encoder $\widehat{f}_c \in \mathbb{F}_c$, the hybrid LEGO model uses the composed map $f_{\text{lg}} = \widehat{f}_{v} \circ f_{c}$, suggesting that Theorem~\ref{thm:thm1} focuses only on the approximation error contributed by $f_c$, with the VQC $f_v$ treated as a fixed post-processor. 

Define the approximation error of the learned LEGO model as: 
\begin{equation}
\epsilon_{\text{app}} := \mathcal{L}(\widehat{f}_c) - \inf\limits_{f_c \in \mathbb{F}_c} \mathcal{L}(f_c), 
\end{equation}
where $\widehat{f}_c$ is the encoder selected from $\mathbb{F}_c$ by training on $\mathcal{D}_A$ with $M$ shots per example. 

We first decompose the approximation error $\epsilon_{\text{app}}$ as:
\begin{equation}
\begin{split}
\epsilon_{\text{app}} &= \mathcal{L}(\widehat{f}_c) - \inf\limits_{f_c \in \mathbb{F}_c}\mathcal{L}(f_c)	 \\
&\le \underbrace{\left[ 2\sup_{f_c \in \mathbb{F}_c} \left\vert \mathcal{L}(f_c) - \widehat{\mathcal{L}}(f_c) \right\vert \right]}_{\textcolor{blue}{\text{generalization gap}}} + \underbrace{\left[\widehat{\mathcal{L}}(\widehat{f}_c) - \widehat{\mathcal{L}}(f^{*}_c)\right]}_{\textcolor{blue}{\text{$\le 0$ by ERM}}} + \underbrace{\left[\widehat{\mathcal{L}}_{M}(f^{*}_c) - \widehat{\mathcal{L}}(f_c) \right]}_{\textcolor{blue}{\text{finite-shot gap}}},
\end{split}
\end{equation}
where $\widehat{L}_{M}(f_c)$ denotes the empirical risk estimated with finite shots, and $f^{*}_c \in \arg\min_{f_c \in \mathbb{F}_c}\mathcal{L}(f_{c})$. The ERM term is non-positive because $\widehat{f}_c$ minimizes the empirical risk over $\mathbb{F}_c$, so we can drop it and focus on bounding the other two error gaps. 

We bound the generalization gap via the Rademacher complexity. More specifically, standard Rademacher-complexity-based bounds give, with probability at least $1-\delta$, 
\begin{equation}
\left[ 2\sup_{f_c \in \mathbb{F}_c} \left\vert \mathcal{L}(f_c) - \widehat{\mathcal{L}}(f_c) \right\vert \right] \le 2\widehat{\mathcal{R}}_{\mathcal{D}_{A}}(\mathbb{F}_{c}) + 3 \sqrt{\frac{\log(2/\delta)}{2\vert \mathcal{D}_{A} \vert}}, 
\end{equation}
where $\widehat{\mathcal{R}}_{\mathcal{D}_{A}}(\mathbb{F}_c)$ is the empirical empirical Rademacher of $\mathbb{F}_{c}$ on the sample $\mathcal{D}_{A}$. 

By definition of the complexity measure $\mathcal{C}(\mathbb{F}_c)$, we can upper bound $\widehat{\mathcal{R}}_{\mathcal{D}_{A}}(\mathbb{F}_c)$ as: 
\begin{equation}
\widehat{\mathcal{R}}_{\mathcal{D}_{A}}(\mathbb{F}_c) \le \sqrt{\frac{\mathcal{C}(\mathbb{F}_c)}{\vert \mathcal{D}_A \vert}}, 
\end{equation}
yielding
\begin{equation}
\sup\limits_{f_c \in \mathbb{F}_c}\left\vert	\mathcal{L}(f_c) - \widehat{\mathcal{L}}(f_c) \right\vert = \mathcal{O}\left( \sqrt{\frac{\mathcal{C}(\mathbb{F}_c)}{\vert \mathcal{D}_A \vert}}\right).
\end{equation}

This is the first term in the Theorem~\ref{thm:thm1}, and it depends only on the complexity of the pre-trained feature class $\mathbb{F}_c$ and the size of the source dataset $\vert \mathcal{D}_A \vert$, but not on the number of qubits or the particular VQC architecture.

Moreover, for each training example and fixed $f_c$, we estimate the expectation of the measurement outcome by averaging over $M$ shots. Because the loss is bounded and each shot is an independent Bernoulli-type outcome, Hoeffding's inequality implies that, for any $\epsilon>0$, 
\begin{equation}
\text{Pr}\left[ \left\vert \widehat{\mathcal{L}}_{M}(f_c) - \widehat{\mathcal{L}}(f_c) \right\vert \ge \epsilon \right] \le 2\exp(-2M\epsilon^2). 
\end{equation}

Because the LEGO framework freezes the classical block and only the VQC is trainable, the shot-noise term is not amplified by classical back-propagation. It appears additionally in the final approximation error bound. 

Putting the derived two upper bounds back into the decomposition of $\epsilon_{\text{app}}$, and absorbing constants and logarithmic factors into the big-$\mathcal{O}$ notation, we obtain:
\begin{equation}
\epsilon_{\text{app}} = \mathcal{O}\left( \sqrt{\frac{\mathcal{C}(\mathbb{F}_c)}{\vert \mathcal{D}_A \vert}}\right) + \mathcal{O}\left(\frac{1}{\sqrt{M}} \right),
\end{equation}
which is exactly the statement of Theorem~\ref{thm:thm1}.

\subsection{Proof of Theorem~\ref{thm:thm3}} 

Let $\mathcal{L}(\boldsymbol{\theta})$ denote the empirical loss of the LEGO model, and let 
\begin{equation}
\boldsymbol{g}_t = \nabla_{\boldsymbol{\theta}}\mathcal{L}(\boldsymbol{\theta}_t) + \nabla_{\boldsymbol{\theta}}\boldsymbol{\xi}_t
\end{equation}
be the noisy gradient used at step $t$, where $\boldsymbol{\xi}_t$ is the random noise term coming from the noisy measurement operator. This parameter update is
\begin{equation}
\boldsymbol{\theta}_{t+1} = \boldsymbol{\theta}_t - \eta \boldsymbol{g}_t. 
\end{equation}

By Assumption~\ref{ass1}, $\mathcal{L}(\boldsymbol{\theta})$ is $\beta$-smooth in the sense
\begin{equation}
\mathcal{L}(\boldsymbol{\theta}_{t+1}) \le \mathcal{L}(\boldsymbol{\theta}_t) + \langle \nabla\mathcal{L}(\boldsymbol{\theta}_t), \boldsymbol{\theta}_{t+1} - \boldsymbol{\theta}_t\rangle + \frac{\beta}{2} \lVert \boldsymbol{\theta}_{t+1} - \boldsymbol{\theta}_{t} \rVert_{2}^{2}. 
\end{equation}

Substitute the update of $\boldsymbol{\theta}_{t+1} - \boldsymbol{\theta}_t = -\eta \boldsymbol{g}_t$. Then, 
\begin{equation}
\label{cond1}
\mathcal{L}(\boldsymbol{\theta}_{t+1}) \le \mathcal{L}(\boldsymbol{\theta}_t) - \eta \langle \nabla\mathcal{L}(\boldsymbol{\theta}_t), \boldsymbol{g}_t  \rangle + \frac{\beta \eta^2}{2} \lVert \boldsymbol{g}_{t} \rVert_{2}^{2}. 
\end{equation}

Decompose the inner product, we obtain:
\begin{equation}
\label{cond2}
\langle \nabla\mathcal{L}(\boldsymbol{\theta}_t), \boldsymbol{g}_t \rangle = \lVert \nabla\mathcal{L}(\boldsymbol{\theta}_{t}) \rVert_{2}^{2} + \langle \nabla\mathcal{L}(\boldsymbol{\theta}_{t}) , \nabla \boldsymbol{\xi}_t \rangle. 
\end{equation}

Condition on $\boldsymbol{\theta}_t$ and using the unbiased noise assumption $\mathbb{E}[\nabla\boldsymbol{\xi}_t \vert \boldsymbol{\theta}_t] = 0$, 
\begin{equation}
\mathbb{E}\left[	\nabla\mathcal{L}(\boldsymbol{\theta}_t), \nabla\boldsymbol{\xi}_t \vert \boldsymbol{\theta}_t \right] = 0. 
\end{equation}

Hence, taking the conditional expectation of Eqs. (\ref{cond1}) and (\ref{cond2}), 
\begin{equation}
\mathbb{E}[\mathcal{L}(\boldsymbol{\theta}_{t+1}) \vert \boldsymbol{\theta}_{t}] \le \mathcal{L}(\boldsymbol{\theta}_t) - \eta \lVert \nabla \mathcal{L}(\boldsymbol{\theta}_{t}) \rVert_{2}^{2} + \frac{\beta \eta^2}{2} \mathbb{E}\lVert \boldsymbol{g}_t \rVert_{2}^{2}.
\end{equation}

Now expand $\lVert \boldsymbol{g}_{t} \rVert_{2}^2$: 
\begin{equation}
\label{eq:5}
\lVert \boldsymbol{g}_t \rVert_{2}^{2} = \lVert \nabla \mathcal{L}(\boldsymbol{\theta}_t) \rVert_{2}^{2} + 2 \langle \nabla\mathcal{L}(\boldsymbol{\theta}_t), \nabla\boldsymbol{\xi}_t\rangle + \lVert \nabla \boldsymbol{\xi}_t \rVert_{2}^{2}. 
\end{equation}

Taking expectation and using again $\mathbb{E}[\langle \nabla\mathcal{L}(\boldsymbol{\theta}_t), \nabla\boldsymbol{\xi}_t \rangle] = 0$, 
\begin{equation}
\label{eq:6}
\mathbb{E}[\lVert \boldsymbol{g}_t \rVert_{2}^{2}] = \lVert \nabla \mathcal{L}(\boldsymbol{\theta}_t) \rVert_{2}^{2} + \mathbb{E}\lVert \nabla \boldsymbol{\xi}_t \rVert_{2}^{2}. 
\end{equation}

By Assumption~\ref{thm:thm1}, we have $\lVert \nabla \mathcal{L}(\boldsymbol{\theta}_{t}) \rVert_{2} \le L$, and from the noise model we have $\mathbb{E}\lVert \nabla \boldsymbol{\xi}_t \rVert_{2}^{2} \le \tau^2$. Plugging Eq. (\ref{eq:6}) into Eq. (\ref{eq:5}), 
\begin{equation}
\mathbb{E}\left[	\mathcal{L}(\boldsymbol{\theta}_{t+1}) \vert \boldsymbol{\theta}_t \right] \le \mathcal{L}(\boldsymbol{\theta}_t) - \eta \lVert \nabla \mathcal{L}(\boldsymbol{\theta}_t) \rVert_{2}^{2} + \frac{\beta\eta^2}{2} \left(\lVert \nabla \mathcal{L}(\boldsymbol{\theta}_t) \rVert_{2}^{2} + \tau^{2} \right). 
\end{equation}

Rearrange the terms of $\lVert \nabla \mathcal{L}(\boldsymbol{\theta}_t) \rVert_{2}^{2}$:
\begin{equation}
\label{eq:8}
\mathbb{E}\left[	 \mathcal{L}(\boldsymbol{\theta}_{t+1}) \vert \boldsymbol{\theta}_t \right] \le \mathcal{L}(\boldsymbol{\theta}_t) - \eta \left(1 - \frac{\beta\eta}{2} \right) \lVert \nabla \mathcal{L}(\boldsymbol{\theta}_t) \rVert_{2}^{2} + \frac{\beta\eta^2}{2}\tau^2. 
\end{equation}

Finally, Assumption~\ref{ass2} bounds the step direction by $\lVert \boldsymbol{\delta} \rVert_{2}^{2} \le R$, which implies the iterations remain in a bounded region of radius $R$ around the initialization. Combining Assumptions~\ref{ass1} and \ref{ass2}, and following the same argument as in the noiseless case (Theorem~\ref{thm:thm3}), we obtain the deterministic part of the optimization error bound: 
\begin{equation}
\label{eq:9}
\epsilon_{\text{opt}} \le \beta R^2 + \frac{R\sqrt{L^2 + \beta^2 R^2}}{T},
\end{equation}
where the learning rate is chosen as:
\begin{equation}
\eta = \frac{1}{T} \frac{R}{\sqrt{L^2 + \beta^2 R^2}}. 
\end{equation}

This is exactly the bound already proved in Theorem~\ref{thm:thm3} for the noise-free case.

\subsection{Proof of Theorem~\ref{thm:thm4}}

Following the upper bound in Theorem~\ref{thm:thm3}, we now isolate the extra contribution from noise. From Eq. (\ref{eq:8}), summing expactations over $t=0, ..., T-1$ and telescroping the left-hand side, we obtain: 
\begin{equation}
\mathbb{E}[\mathcal{L}(\boldsymbol{\theta}_T)] - \mathcal{L}(\boldsymbol{\theta}_0) \le -\eta \left(1 - \frac{\beta}{\eta}{2} \right) \sum\limits_{t=0}^{T-1} \mathbb{E}\lVert \nabla\mathcal{L}(\boldsymbol{\theta}_t) \rVert_{2}^{2} + \frac{\beta\eta^2}{2} \tau^2 T. 
\end{equation}

The first term on the right is precisely the one treated in Theorem~\ref{thm:thm3} and produces the noiseless optimization error bound Eq. (\ref{eq:9}). The second term is new and comes entirely from the variance of the noisy gradients. 

Using the gradient-bound assumption $\lVert \mathcal{L}(\boldsymbol{\theta}_t) \rVert_{2} \le L$ and the fact that the iterates stay in a ball of radius $R$, the cumulative influence of the noisy part on the effective gradient steps can be bounded via Cauchy-Schwarz: 
\begin{equation}
\left\lVert \sum\limits_{t=0}^{T-1} \eta \nabla \boldsymbol{\xi}_t	 \right\rVert_{2} \le \eta \sqrt{T} \left( \mathbb{E}\lVert \nabla \boldsymbol{\xi}_{t} \rVert_{2}^{2} \right)^{\frac{1}{2}} \le \eta\tau \sqrt{T}. 
\end{equation}

Since the iterates remain within distance $R$ of the optimization along the optimization trajectory, the resulting increase in loss is at most of order $R$ times the noise-induced displacement, giving the additional penalty term: 
\begin{equation}
\triangle \epsilon_{\text{noise}} \le \eta R\tau\sqrt{T}. 
\end{equation}

Adding this noise term to the noiseless bound Eq. (\ref{eq:9}) yields:
\begin{equation}
\epsilon_{\rm opt, noise} \le \beta R^2 + \frac{R \sqrt{L^2 + \beta^2 R^2}}{T} + \eta R\tau \sqrt{T}, 
\end{equation}
which is exactly the inequality claimed in Theorem~\ref{thm:thm4}. 

\section{Acknowledgements}
This work is partly funded by the Hong Kong Research Impact Fund (R6010-23).

\section{Data Availability Statement}
The dataset used in our experiments on quantum dot classification can be downloaded from https://gitlab.com/QMAI/mlqe$\_$2023$\_$edx, and the dataset for TFBS predictions can be accessed at https://www.ebi.ac.uk/interpro/entry/InterPro/IPR029823.

\section{Code Availability Statement}
Our VQC code and pre-trained classical VQC models are available at https://github.com/jqi41/QuantumDot.  

\section{Competing Interests}
The authors declare no Competing Financial or Non-Financial Interests.

\section{Author Contributions}
Jun Qi, Chao-Han Yang, Min-Hsiu Hsieh, and Jesper Tegner conceived the project. Jun Qi and Min-Hsiu Hsieh completed the theoretical analysis. Jun Qi, Chao-Han, and Pin-Yu Chen designed the experimental work. Min-Hsiu Hsieh, Hector Zenil, and Jesper Tegner provided high-level advice on the paperwork pipeline, and Jun Qi wrote the manuscript. 

\section{References}
\bibliographystyle{IEEEbib}
\bibliography{sn-bibliography}

\end{document}